\documentclass{IOS-Book-Article}

\usepackage{mathptmx}
\usepackage{soul}\setuldepth{article}
\usepackage{graphicx}

\def\hb{\hbox to 11.5 cm{}}

\begin{document}

\pagestyle{headings}
\def\thepage{}

\begin{frontmatter}              

\title{A Deep Learning-Based System for\\ Automatic Case Summarization}
\author[A]{Minh Duong}
\author[A]{Long Nguyen}
\author[A]{Yen Vuong}
\author[A]{Trong Le}
\author[C]{Ha-Thanh Nguyen}

\address[A]{Vietnam National University - Hanoi, Vietnam}
\address[C]{National Institute of Informatics, Japan}




\begin{abstract}
This paper presents a deep learning-based system for efficient automatic case summarization. Leveraging state-of-the-art natural language processing techniques, the system offers both supervised and unsupervised methods to generate concise and relevant summaries of lengthy legal case documents. The user-friendly interface allows users to browse the system's database of legal case documents, select their desired case, and choose their preferred summarization method. The system generates comprehensive summaries for each subsection of the legal text as well as an overall summary. This demo streamlines legal case document analysis, potentially benefiting legal professionals by reducing workload and increasing efficiency. Future work will focus on refining summarization techniques and exploring the application of our methods to other types of legal texts.

\end{abstract}

\end{frontmatter}

\section{Introduction}

Analyzing legal case documents is a critical and labor-intensive aspect of the legal profession in Vietnam. Lawyers, judges, and other legal professionals are required to examine vast amounts of intricate texts \cite{nguyen2017knowledge} to comprehend cases, detect legal issues, and arrive at informed decisions. There has been extensive research in the domain of text summarization \cite{huang2020have}, and some methodologies have been uniquely adapted for legal text summarization \cite{polsley2016casesummarizer, zhong2019automatic, farzindar2004legal}. To further support the Vietnamese legal community, and potentially other similar legal systems worldwide, we propose an automatic legal case document summarization system. This system utilizes cutting-edge natural language processing techniques for extractive summarization, providing users with concise and relevant summaries of lengthy Vietnamese legal case documents, effectively facilitating their legal analysis. This demo paper initially offers a brief background on legal document summarization specific to the Vietnamese context, then outlines the overall architecture and the user interface of our system.

\section{Background}

Text summarization has been a popular research area in natural language processing, with various techniques being proposed and evaluated over the years, such as TextRank \cite{mihalcea2004textrank}, LexRank \cite{erkan2004lexrank}, and sentence reduction \cite{jing2000sentence}. The development of pretrained language models like BERT \cite{devlin2018bert}, GPT-2 \cite{radford2019language}, BART \cite{lewis2019bart}, and T5 \cite{raffel2020exploring} has further advanced the state of text summarization, including both extractive and abstractive approaches \cite{liu2019text, zhang2020pegasus}.

Legal document summarization presents unique challenges, as legal texts often contain complex syntactic structures, domain-specific terminology, and arguments requiring deep understanding of the content \cite{farzindar2004legal, saravanan2006improving}. Several approaches have been developed to tackle legal text summarization \cite{polsley2016casesummarizer, zhong2019automatic}, with some focusing on the specific structure and argumentative roles of legal documents \cite{farzindar2004legal, feijo2023improving}. Other researchers have applied deep learning techniques for related legal tasks, such as case law retrieval \cite{vuong2022sm,nguyen2022attentive} and legal question answering \cite{kien2020answering}. Given these advances and the importance of legal document analysis, there is great potential in developing an automatic system for legal case document summarization.

\section{System Description}

The primary goal of our automatic legal case document summarization system is to provide users with a concise and relevant summary of lengthy legal documents. Our system can be divided into two main components: the general architecture and the user interface.

\subsection{General Architecture}

Our system harnesses state-of-the-art natural language processing techniques to summarize case law documents. This includes two distinct approaches: the unsupervised and supervised methods. Both methods extract key information from the document and present the most relevant sentences as the summary. The unsupervised method is based on graph-based calculations, such as TextRank \cite{mihalcea2004textrank}, and the supervised method employs a fine-tuned multilingual BERT model \cite{devlin2018bert} that has been trained on a large dataset of Vietnamese legal cases.

\begin{figure}
\label{fig:genarch}
\includegraphics[width=.6\textwidth]{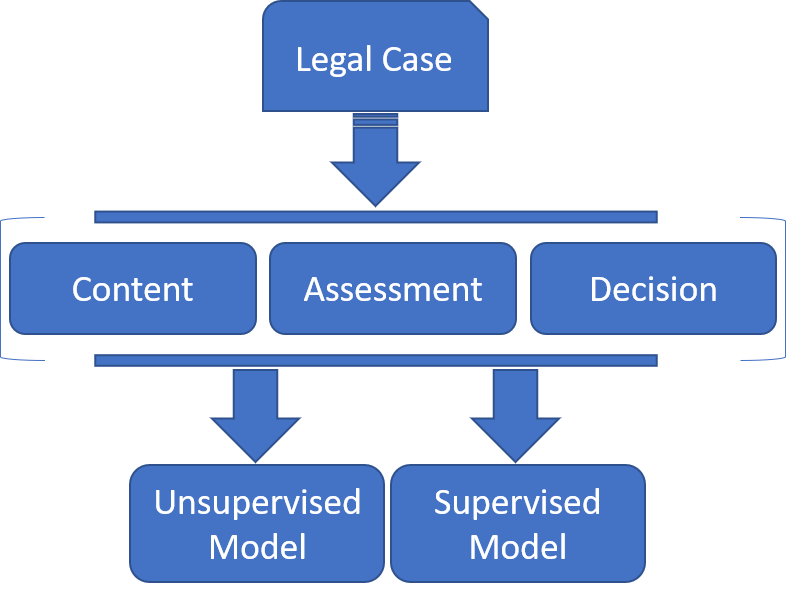}
\centering
\caption{General Architecture of the System}
\end{figure}

The system first segments legal texts into subsections and then employs either the supervised or unsupervised summarization method, depending on the user's preference, to generate summaries for each subsection. Finally, summaries of all subsections are combined to create a comprehensive summary of the entire legal case document.
According to Article 266 of the Civil Procedure Code 2015\footnote{https://wipolex-res.wipo.int/edocs/lexdocs/laws/en/vn/vn083en.html}, the structure of a court judgment consists of three parts: the Introduction, the Case Content and Assessment, and the Court's Decision. The main contents of a summary generally lie in the three sections: Content, Assessment, and Decision. Figure \ref{fig:genarch} illustrates the general architecture of the system.

\subsection{User Interface}

\begin{figure}
\label{fig:ui}
\includegraphics[width=.9\textwidth]{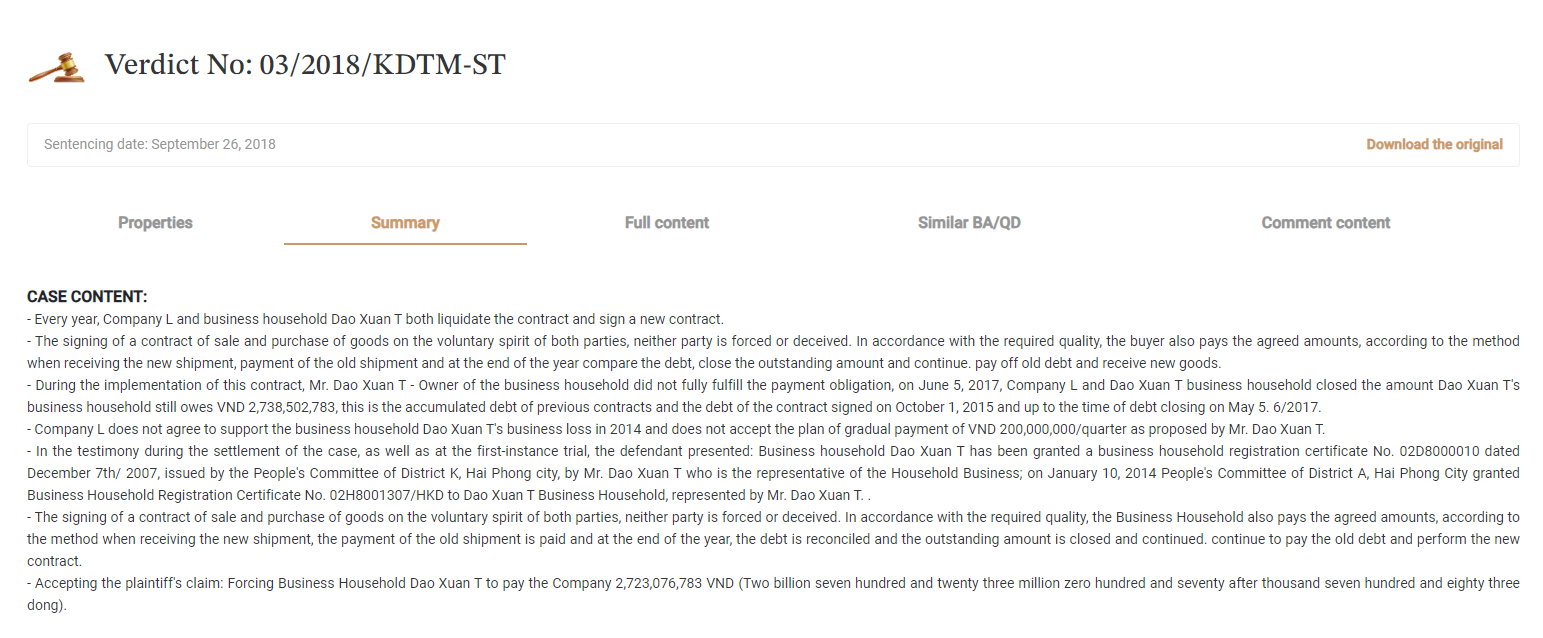}
\centering
\caption{The summarization module integrated into the system's user interface.}
\end{figure}

Our system is designed with an intuitive user interface, making it easy for non-experts to understand and navigate. Users are provided with an option to browse the system's database of legal case documents, organized by categories such as case subject matter, jurisdiction, and date of decision. After the user selects a desired case, the system automatically applies a mechanism to choose the best summary among the ones generated by both supervised and unsupervised summarization methods. Our system provides a full-text summary for users, divided into sections such as Content, Assessment, and Decision, with bullet points for easy tracking, comparison, and analysis. The summarized results are displayed directly on the screen for users to view (See Figure \ref{fig:ui}).

\section{Conclusions}

In this demo paper, we have presented an automatic legal case document summarization system utilizing both unsupervised and supervised methods to generate concise and relevant summaries. We provided a brief background on legal document summarization and described the overall architecture of the system and the user interface, which allows users to browse a comprehensive database of legal cases. Our system marks a significant step towards streamlining legal case analysis and has the potential to benefit legal professionals in their daily work greatly. In future work, we plan to refine and expand upon the summarization methods employed and explore the application of our techniques to other types of legal texts.

\section*{Acknowledgements}

The authors would like to express their gratitude to LexEngine JSC Vietnam for their valuable collaboration in implementing the expert-driven labelling and the computational platform. Their support and expertise greatly contributed to the successful completion of this research and the development of the presented paper.

\bibliographystyle{vancouver}
\bibliography{ref.bib}




\end{document}